\documentclass[letterpaper, 10 pt, conference]{ieeeconf}  % Comment this line out if you need a4paper
\usepackage{amssymb}
\usepackage{amsmath}
\usepackage{graphicx}
\usepackage{float} 

\IEEEoverridecommandlockouts                              % This command is only needed if 
\title{\LARGE \bf
Categorical Policies: Multimodal Policy Learning and Exploration in Continuous Control
}

\author{SM Mazharul Islam$^{\dagger~1}$ and Manfred Huber$^{1}$% <-this % stops a space
\thanks{$^{\dagger}$ corresponding author, {\tt\small mazharul2752@gmail.com}}% <-this % stops a space
\thanks{$^{1}$ are with the Department of Computer Science and Engineering, University of Texas at Arlington, TX-76019, USA}
}

\begin{document}

\maketitle
\thispagestyle{empty}
\pagestyle{empty}

%%%%%%%%%%%%%%%%%%%%%%%%%%%%%%%%%%%%%%%%%%%%%%%%%%%%%%%%%%%%%%%%%%%%%%%%%%%%%%%%
\begin{abstract}
A policy in deep reinforcement learning (RL), either deterministic or stochastic, is commonly parameterized as a Gaussian distribution alone, limiting the learned behavior to be unimodal. However, the nature of many practical decision-making problems favors a multimodal policy that facilitates robust exploration of the environment and thus to address learning challenges arising from sparse rewards, complex dynamics, or the need for strategic adaptation to varying contexts. This issue is exacerbated in continuous control domains where exploration usually takes place in the vicinity of the predicted optimal action, either through an additive Gaussian noise or the sampling process of a stochastic policy. In this paper, we introduce Categorical Policies to model multimodal behavior modes with an intermediate categorical distribution, and then generate output action that is conditioned on the sampled mode. We explore two sampling schemes that ensure differentiable discrete latent structure while maintaining efficient gradient-based optimization. By utilizing a latent categorical distribution to select the behavior mode, our approach naturally expresses multimodality while remaining fully differentiable via the sampling tricks. We evaluate our multimodal policy on a set of DeepMind Control Suite environments, demonstrating that through better exploration, our learned policies converge faster and outperform standard Gaussian policies. Our results indicate that the Categorical distribution serves as a powerful tool for structured exploration and multimodal behavior representation in continuous control.

\end{abstract}

%%%%%%%%%%%%%%%%%%%%%%%%%%%%%%%%%%%%%%%%%%%%%%%%%%%%%%%%%%%%%%%%%%%%%%%%%%%%%%%%
\section{INTRODUCTION}

Reinforcement learning (RL) in general has made remarkable strides in recent years, matching or surpassing human base-line performance in diverse fields \cite{haarnoja2018soft, baker2022video, chen2024end, hafner2020mastering}. Learning accurate dynamics of the environment within a compact latent space \cite{amos2018learning, hafner2019learning} has significantly enhanced the generalization of the high-dimensional state space, leading to drastically improved sample efficiency with model-based RL methods. The generative nature of such latent-space models enables agents to generate large numbers of action-conditioned imagined trajectories to learn an optimal policy with minimal computational overhead.

Traditional model-free algorithms, such as policy gradient \cite{sutton1999policy}, actor-critic \cite{konda1999actor}, proximal policy optimization \cite{schulman2017proximal}, and their variants \cite{silver2014deterministic, haarnoja2018soft}, can be seamlessly integrated with such latent-space models for long-horizon planning, as thousands of imagined trajectories can be simulated in parallel \cite{hafner2019learning}. As a result, recent model-based algorithms \cite{hafner2019dream, hafner2020mastering} have demonstrated notable improvements in both raw performance and training efficiency across popular benchmark environments, such as the DeepMind Control Suite \cite{tassa2018deepmind}, Atari 100k \cite{mnih2015human}, Minecraft \cite{hafner2023mastering}, Super Mario \cite{pathak2017curiosity} etc.

The learned policy itself is either deterministic \cite{hafner2019learning}, incorporating predefined action noise for exploration, or stochastic \cite{haarnoja2018soft}, with entropy regularization to encourage exploration. In continuous control, such policies are often parameterized as a multivariate diagonal Gaussian distribution \cite{lillicrap2015continuous}. This approach has the benefit of enabling efficient gradient-based optimization, ensuring smooth and consistent action selection, and simplified learning by assuming independence between action dimensions. A key limitation with such approaches is that the policy tends to optimize for task completion greedily, ignoring to maintain and explore a rich repertoire of behaviors. This often results in sub-optimal policies that are narrowly focused, lacking the ability to generalize across varying contexts. Hence, the standard exploration-exploitation setup along with a Gaussian policy often collapses to a single mode, preventing an agent from learning a diverse, multimodal behavior that is usually observed in biological agents.

\begin{figure*}[ht]
      \centering
      % \framebox{\parbox{3in}{We suggest that you use a text box to insert a graphic (which is ideally a 300 dpi TIFF or EPS file, with all fonts embedded) because, in an document, this method is somewhat more stable than directly inserting a picture.
% }}
      \includegraphics[width=0.95\linewidth]{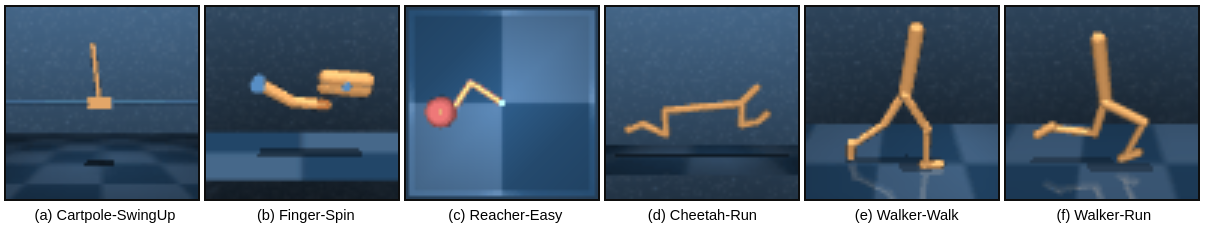}
      \caption{Agent observations after raw frames are rescaled to $3 \times 64 \times 64$ pixels. Subfigures (a)–(c) correspond to environments with smaller action spaces. (a) has a single action dimension, while (b) and (c) each have two. In contrast, (d)–(f) represent environments with larger action spaces, each comprising six action dimensions.}
      \label{figurelabel}
   \end{figure*}

A multimodal behavior policy offers several key advantages over a unimodal one. A unimodal policy, by definition, provides a single action prediction—typically the mean of the distribution—for a given state. This can be limiting in scenarios where the predicted action becomes infeasible, potentially causing the policy to fail entirely. For example, consider an agent tasked with making a cup of coffee from raw ingredients, finds liquid milk has run out, and now to succeed must use powdered milk instead. If the agent expects to use liquid milk but finds it is unavailable, a unimodal policy may struggle to adapt. Although it may have learned to use either liquid milk or powdered milk, it cannot flexibly choose between them at test time. In contrast, a multimodal policy can represent multiple viable behaviors, enabling the agent to switch strategies seamlessly. Moreover, multimodal policies can more readily incorporate new tasks or alternative solutions without requiring the agent to \emph{unlearn} previously acquired behaviors, naturally promoting structured and diverse exploration even after convergence, which is essential for long-term adaptability and robustness.

In this paper, instead of relying on a Gaussian distribution alone, we integrate intermediate categorical distribution to select a discrete behavior mode first and then condition the output action on the discrete behavior mode. A single categorical variable struggles to scale effectively, as achieving fine-grained control would require an impractically large number of classes to represent all possible behavior modes. Hence, we use multiple categorical variables with fewer classes each to form a combinatorial representation of behaviors. This not only reduces the total number of parameters needed but also provides a more structured and expressive policy space, enabling the agent to capture a wider variety of behaviors and adapt more effectively to complex tasks. Further, since a naive sampling process disconnects the computational graph after the sampling, we experiment with straight-through estimation (STE) \cite{bengio2013estimating} and the Gumbel-Softmax reparameterization trick \cite{maddison2016concrete} as ways to retain a fully differentiable computational graph. Gumbel-Softmax does not introduce any bias to the policy with the cost of higher variance attributed to the added sampling process after Softmax. On the other hand, though STE introduces some bias through the approximation, the estimated gradient has low variance due to the absence of sampling. As a result, our method is straightforward to implement using commonly available automatic differentiation libraries \cite{paszke2019pytorch}. To demonstrate its effectiveness, we evaluate categorical policies on a set of continuous control tasks selected from the DeepMind Control Suite \cite{tassa2018deepmind} environment shown in Figure 1. In order to reduce the training time, we utilize commonly available dynamics models \cite{ha2018world, hafner2019learning, hafner2019dream} and train our policies within the framework of model-based RL. However, it is important to note that our approach remains agnostic to both model-free and model-based RL methods, making it broadly applicable and easy to integrate with existing solutions.

The main contributions of this article include:
\begin{itemize}
    \item Introducing a novel approach to parameterize a multimodal policy through an intermediate categorical distribution.
    \item Empirical evaluation of two methods, namely STE and Gumbel-Softmax reparameterization trick, that allows the gradient to flow past the discrete sampling step.
\end{itemize}

The remainder of the paper is organized as follows. Section \ref{rel_work} provides a review of the relevant literature. In Section \ref{prelim}, we introduce the foundational concepts in RL and the Concrete distribution. Section \ref{multi_modal} presents a detailed formulation of multimodal policy. In Section \ref{exp}, we present the experimental findings that support our claims. Finally, Section \ref{concl} concludes the paper and discusses potential future directions.

\section{RELATED WORKS}
\label{rel_work}
Recent advances in both online \cite{haarnoja2018soft, barth2018distributed} and offline RL \cite{kumar2020conservative, levine2020offline} have led to significant improvements in learning policies for continuous control tasks. Successfully modeling the environment within a latent space has further pushed this boundary to high-dimensional visual observation based environments. Despite such progress on the model side, most current implementations rely on simple deterministic or stochastic policies, typically conditioned on the approximated states using a Gaussian distribution \cite{lillicrap2015continuous}, which restricts the learned behavior to a single mode. Goal-oriented RL \cite{bacon2017option, chane2021goal}  methods have attempted to address this limitation by conditioning policies on various goals. However, these methods still face the challenge of unimodal behavior modeling, as the policy is typically optimized for one goal at a time, limiting the diversity of the learned behavior for a given state.

Discovering skills \cite{konidaris2009skill} from either demonstrated data \cite{shankar2020learning} or unsupervised RL \cite{eysenbach2018diversity} is a promising direction in the context of multimodal behavior learning. A common approach involves learning a diverse set of lower-level policies \cite{kim2023variational, mendonca2021discovering} that can be stitched together to perform complex tasks. However, these methods often require additional mechanisms to generate intermediate goals or design complex reward functions, which can introduce significant engineering overhead and limit scalability.

Hierarchical RL methods \cite{bacon2017option, achiam2018variational}, such as option learning, aim to address these challenges by enabling the discovery of state-specific action sequences with well-defined initiation conditions and optional termination criteria. While these structured policies promise improved interpretability and modularity, they come with notable trade-offs. The added complexity in action inference and planning can lead to higher computational costs during both training and inference. Moreover, despite the hierarchical structure, the final policy often remains a greedy optimization of individual skills or subgoals, leading to limited behavioral diversity and reduced adaptability in dynamic environments. 

Another common direction to achieving multimodal behavior is the discretization of continuous action spaces into a binary format \cite{sallans2004reinforcement}, allowing policy optimization within Markov Decision Processes (MDPs) with large action sets. While this method can be effective for tasks with low-dimensional action spaces, it quickly becomes infeasible in higher dimensions due to the curse of dimensionality. The exponential growth of discrete action combinations leads to inefficient exploration and poor scalability, limiting its practicality for complex tasks. To mitigate this, some methods assume independence between action dimensions, effectively factorizing the action space. While this assumption reduces computational complexity, it fails to capture the inherent structure and dependencies present in continuous spaces \cite{tavakoli2018action}, leading to suboptimal performance. More advanced approaches address this limitation by introducing a factorized distribution across action dimensions that preserves the ordinal relationships within the discrete representation \cite{tang2020discretizing}. However, these methods often come at the cost of high variance in gradient estimation and instability during training, making learning less reliable.

\section{PRELIMINARIES}
\label{prelim}

\subsection{Reinforcement Learning}
We consider a partially observable MDP (POMDP) with discrete time steps $t \in \{1, \dots, T\}$. At each $t$ an agent receives a high-dimensional visual observation $o_t \in \mathbb{R}^{c \times h \times w}$, performs a $k$-dimensional continuous action $a_t \in \mathbb{R}^k$ generated by some policy $a_t \sim \pi(a_t|o_{1:t}, a_{1:t-1}) $, and receives a scalar reward $r_t \in \mathbb{R}$ and the next observation $o_{t+1}$, according to the unknown environment dynamics $o_t, r_t \sim p(o_t, r_t, d_t | o_{1:t}, a_{1:t-1})$.

The goal of an RL algorithm is to find a policy $\pi$ that maximizes the expected sum of discounted rewards $\mathbb{E}\big[\sum_{t=0}^{\infty}\gamma^{t-1}r_t\big]$, where $\gamma \in [0, 1)$ is the discount factor.

In the context of model-based RL algorithms, with a convolutional neural network (CNN)-based encoder \cite{ha2018world, hafner2019dream}, visual observation $o_t$ is compressed into a continuous vector-valued latent state $s_t$ using a representation model $\mathcal{F}_{rep}$ that satisfies the Markovian transition \cite{watter2015embed}. A transition model $\mathcal{F}_{d}$ predicts the next latent state $s_{t+1}$ without access to the corresponding visual observation $o_{t+1}$ and a reward model $\mathcal{F}_r$ predicts the reward given the current latent state.

\begin{table}[h]
    \centering
    \normalsize{
    \begin{tabular}{c c}
        representation model: &  $\mathcal{F}_{rep}(s_t | s_{t-1}, a_{t-1}, o_t)$ \\
        transition model: & $\mathcal{F}_{d}(s_t | s_{t-1}, a_{t-1})$ \\
        reward model: & $\mathcal{F}_r(r_t | s_t)$ \\
        % policy: & $\pi(a_t | s_t)$ \\
        latent-state: & $s_t = h_t \oplus h^s_t$ \\
    \end{tabular}
    % \caption{Caption}
    \label{tab:dreamer}
    }
\end{table}

Latent state $s_t$ is usually a concatenation between the deterministic hidden state $h_t$ of the sequential model and a stochastic state $h^s_t$ obtained from $h_t$ to address the inherent uncertainty of the environment.

% The policy is parameterized by a diagonal Gaussian distribution, and is trained with standard actor-critic loss.

\subsection{Straight-Through Estimation}
STE \cite{bengio2013estimating}, through a surrogate gradient during backpropagation, offers a simple but effective solution to overcome the challenge due to the non-differentiability of categorical sampling. Given a discrete variable $z$ obtained through $argmax$ or rounding, STE approximates the gradient as if the operation were the identity function:

\begin{equation*} z = \text{argmax}(\mathbf{p}) \quad \text{(forward)}, \hspace{1em}
% \end{equation*} 
% \vspace{0.1em}
% \begin{equation*} 
\frac{\partial z}{\partial \mathbf{p}} \approx \frac{\partial \mathbf{p}}{\partial \mathbf{p}} = \mathbf{I} \quad \text{(backward)}. 
\end{equation*} 

Effectively, STE treats the discrete operation as differentiable in the backward pass, allowing gradients to flow through non-differentiable functions.

\subsection{Gumbel-Softmax Reparameterization}
Gumbel-Softmax reparameterization \cite{maddison2016concrete} provides a differentiable approximation of categorical sampling through a continuous relaxation of the categorical distribution. Given logits $\l = (\l_1, \l_2, \dots, \l_k)$ of a categorical distribution and a temperature parameter $\lambda > 0$, a discrete random variable $\mathbf{z} \in \Delta^{k-1}$ (the $(k-1)$-dimensional probability simplex) is sampled as:
\begin{equation}
z_i = \frac{\exp((\l_i + g_i) / \lambda)}{\sum_{j=1}^{k} \exp((\l_j + g_j) / \lambda)}, \quad \text{for } i = 1, \dots, k,
\label{eq:concrete_dist}
\end{equation}

where $g_i$ are independent samples drawn from the Gumbel distribution: $g_i \sim \text{Gumbel}(0,1)$. As $\lambda \to 0$, samples from the concrete distribution $z$ approximate a categorical sample, while larger $\lambda$ values result in smoother and more uniform probability distributions.

\section{MULTIMODAL POLICY WITH CATEGORICAL DISTRIBUTION}
\label{multi_modal}
In this section, we formulate a multimodal policy $\pi^\mu$ for continuous control using an intermediate categorical distribution coupled with differentiable discrete sampling. Our multimodal policy consists of two stages. First, we sample a behavior mode $b_t$ from the latent state $s_t$ as a categorical variable, allowing the policy to encode a multimodal latent space through a discrete representation. Then, the output action $a_t$ is conditioned on this discrete behavior mode and parameterized by the standard diagonal Gaussian. This hierarchical structure enables the policy to capture complex multi-modal behaviors, improving expressiveness and allowing more structured exploration.

\begin{equation}
\pi^\mu \left\{
\begin{array}{ll}
    b_t \sim \text{Categorical}\big(\mathcal{F}_b(s_t)\big) \\
    a_t \sim \text{Diagonal-Gaussian}\big(\mathcal{F}_a(b_t)\big)
\end{array}
\right.
\label{eq:multimodal_policy}
\end{equation}

Equation \ref{eq:multimodal_policy} abstracts this two-stage process, where $\mathcal{F}$ represents any arbitrary function approximation parameterized by multi-layer perceptron (MLP). The mean and variance of the diagonal Gaussian are obtained by splitting the output of the second MLP into two. For finite range action space (e.g. in DMC environments $\mathcal{A} = [-1, 1]^k$), a tanh-truncated Gaussian distribution is used instead. Resulting policy $\pi^{\mu}$ is trained with actor-critic framework \cite{konda1999actor} described below, where $v_{\lambda}$ is an exponentially-weighted average \cite{hafner2019dream} and $H$ is the imagination horizon.

\begin{table}[h]
    \centering
    \normalsize{
    \begin{tabular}{c c}
        Action model: &  $a_t \sim \mathcal{F}_a(b_t), \quad b_t \sim \mathcal{F}_b(s_t)$ \\
        Value model: & $v_t = v(s_t) = \mathcal{F}_v(s_t) \approx \sum_{t=\tau}^{\tau+H} \gamma^{t - \tau} r_t$ \\
        Action loss: & $ \displaystyle \max_{\mathcal{F}_a, \mathcal{F}_b} E \Big( \sum_{t=\tau}^{\tau+H} v_{\lambda}(s_t)\Big)$ \\
        % policy: & $\pi(a_t | s_t)$ \\
        Value loss: & $ \displaystyle \min_{\mathcal{F}_v} E \Big( \sum_{t=\tau}^{\tau+H} \frac{1}{2} || v(s_t) - v_{\lambda}(s_t)||^2\Big) $
    \end{tabular}
    % \caption{Caption}
    % \label{tab:dreamer}
    }
\end{table}

\begin{figure}[h]
    \centering
    \includegraphics[width=0.6\linewidth]{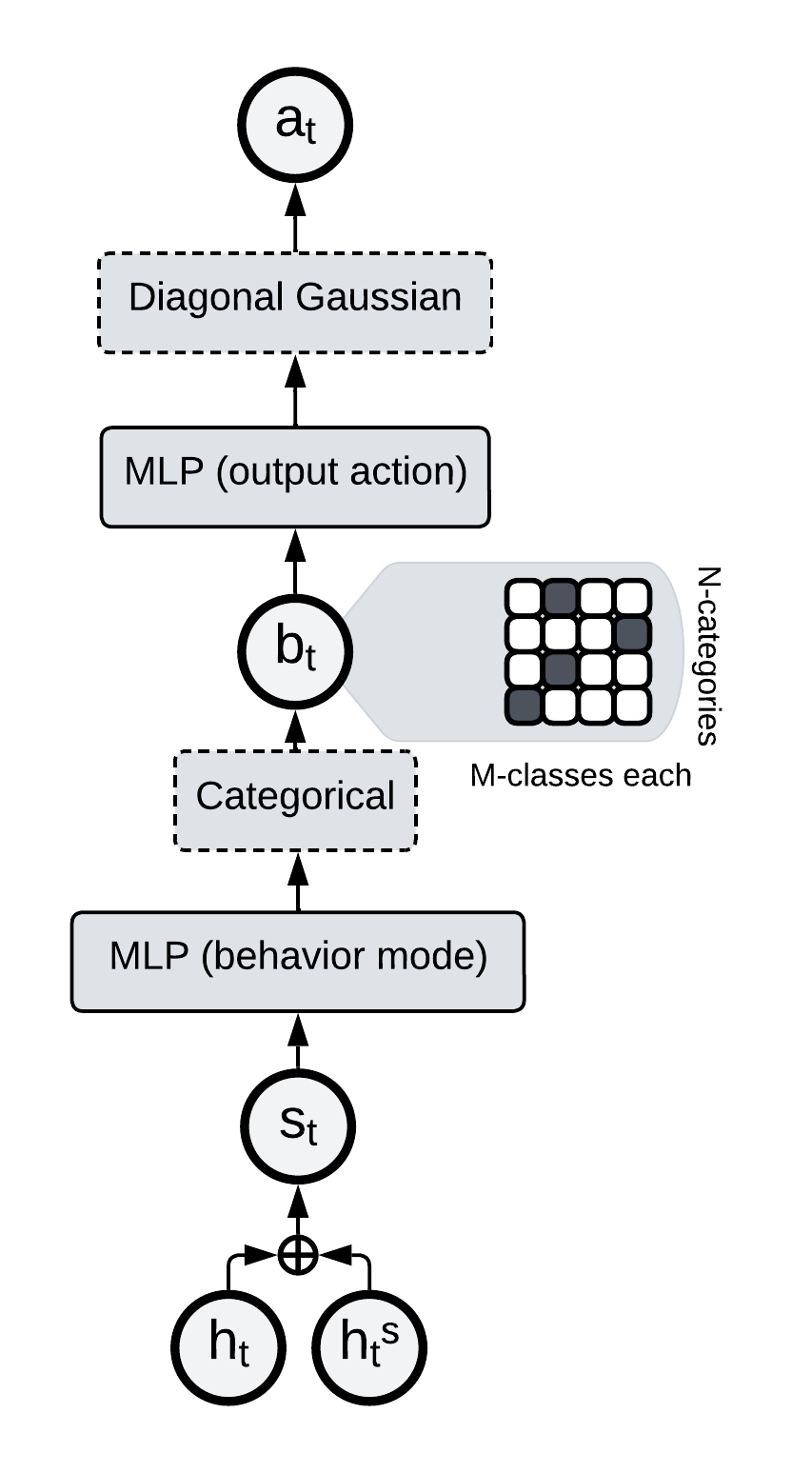}
    % \captionsetup{justification=centerlast, singlelinecheck=false}
    \caption{Multimodal Policy with categorical distribution. Dashed boxes represent sampling step from a distribution. Instead of a single categorical, we opt for a vector of multiple categorical variables to represent the behavior mode.}
    \label{fig:concrete_policy}
\end{figure}

To parameterize $b_t$, one might opt for a single categorical variable with $M$ classes, where each class represents a distinct behavior mode. However, for successful continuous control, the policy must be capable of generating fine-grained continuous actions. But since we desire the output action $a_t$ to be conditioned solely on behavior mode $b_t$, without direct access to the state $s_t$, a prohibitively large number of discrete modes would be required to represent the optimal action distribution. A more structured alternative is to introduce $N$ categorical variables, each with $M$ classes. This approach provides a compositional representation, where the total number of discrete modes is $M^N$, allowing for a sparse yet expressive encoding of diverse behaviors. By factorizing the latent space into multiple categorical components, the policy can efficiently capture complex variations in action modes while keeping the number of discrete choices manageable. We present a visual representation of our multimodal policy framework in Figure \ref{fig:concrete_policy}.

Obtaining behavior mode $b_t$ involves discrete sampling and a naïve approach here would detach the computational graph, preventing gradients from flowing through. One widely-used solution to this problem is to use STE \cite{bengio2013estimating}, where the categorical variable is sampled discretely during the forward pass, but during backpropagation \cite{rumelhart1986learning}, the gradients are approximated from the pre-sampling logits. An alternative is to use Gumbel-Softmax reparameterization which approximates the categorical distribution itself with a continuous relaxation that is differentiable. Instead of directly sampling from a categorical distribution, the Gumbel-Softmax method draws from a softened categorical distribution, producing a smooth one-hot vector. This allows the gradients to flow more smoothly through the continuous relaxation, which can help mitigate the high variance and bias issues of STE. Additionally, the Gumbel-Softmax provides more control over the hardness of the categorical choice by adjusting a temperature parameter, which allows for a continuous trade-off between a soft distribution and the discrete categorical distribution as training progresses. We experiment with both approaches and present comparative results in the next section.

\section{EXPERIMENTS}
\label{exp}

Since our proposed multimodal policy is orthogonal to most existing algorithms across both online and offline reinforcement learning paradigms, we choose a model-based framework \cite{hafner2019learning, hafner2019dream} to learn the environment dynamics. The online nature of this approach enables us to rigorously evaluate the exploration strategies induced by different policy designs. We follow the hyperparameters from the original papers, with one key modification: instead of collecting entire episodes before training, we update all modules every 8 environment steps, as suggested by \cite{hafner2020mastering}. Our multimodal policy network comprises two multilayer perceptrons ($\mathcal{F}_a$ and $\mathcal{F}_b$), each with two hidden layers of 300 units and ELU activations. We do not use dropout or normalization layers. For discrete mode selection with Gumbel-Softmax sampling, we use fixed temperature $\tau = 2.0$ and employ the hard-sampling variant. Although soft-sampling provides smoother gradient flow through the discrete latent space, it also necessitates temperature annealing to align training-time and evaluation-time behavior modes. We opt for hard-sampling to reduce complexity and ensure consistent behavior throughout training and testing. The baseline unimodal policy is implemented as a MLP with three hidden layers of 300 units and ELU activations.

We evaluated our approach on the six continuous control tasks from the DeepMind Control Suite \cite{tassa2018deepmind} shown in Figure 1, chosen to benchmark and compare the performance of unimodal and multimodal policies. These tasks collectively present diverse challenges: long-horizon planning and occlusion memory (Cartpole), sparse rewards (Reacher), high-dimensional state and action spaces (Cheetah), and sequential decision-making (Walker), among others. All models are trained on a single NVIDIA A6000 GPU for $10^7$ environment steps, with each run taking approximately 14 hours of wall-clock time. We store the replay buffer directly in GPU memory to reduce data transfer overhead between CPU and GPU, enabling faster training iterations. For each task, we train three policy variants: a unimodal baseline, a multimodal policy with STE, and a multimodal policy using Gumbel-Softmax sampling.

\subsection{Multimodal versus Unimodal}
\begin{figure}[h]
      \centering
      \includegraphics[width=0.99\linewidth]{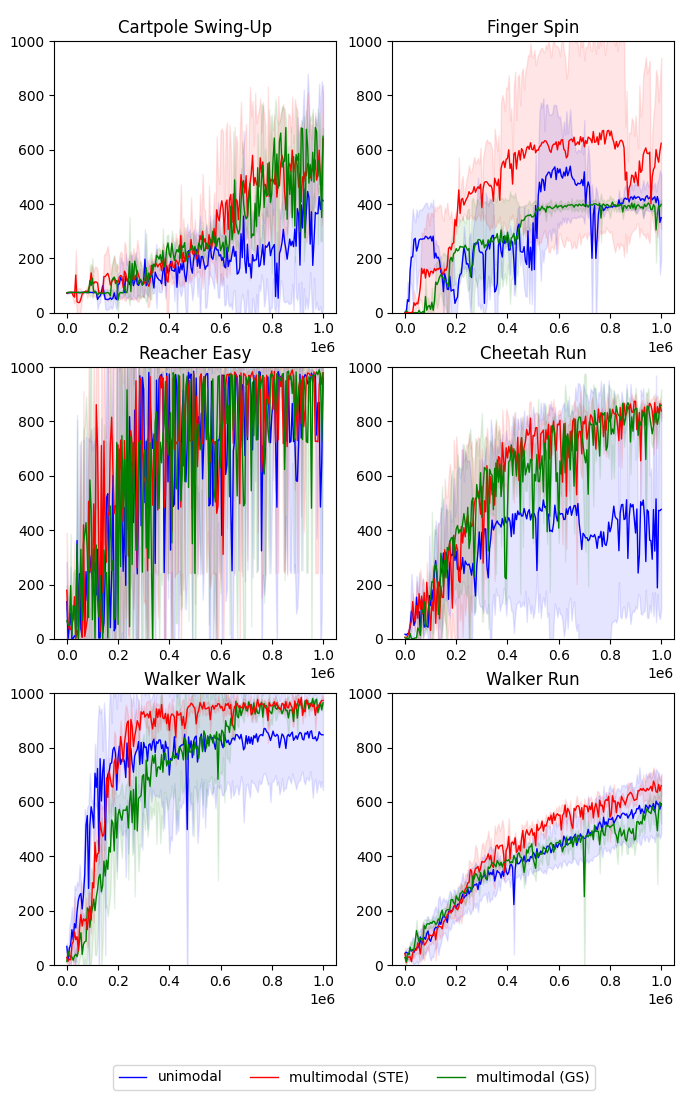}
      \caption{Comparison of multimodal policies against a standard unimodal policy. Plot shows test performance over environment steps. The line shows the average and the shaded area represents 1 standard deviation over 5 seeds.}
      \label{fig:exp_0}
\end{figure}

Figure \ref{fig:exp_0} presents the test-time performance of our proposed multimodal policy on the six tasks and comparison with the baseline unimodal policy. Our results demonstrate that the multimodal policy consistently matches or outperforms the unimodal counterpart in terms of faster convergence, higher episode rewards, and improved robustness (i.e., lower variance across different random seeds).

We attribute the superior performance of the multimodal policy to its structured exploration mechanism, which allows efficient navigation of the action space by leveraging multiple behavior modes. Unlike unimodal policies, which often struggle with mode collapse and may become stuck in suboptimal behaviors, our multimodal approach facilitates more diverse action strategies, leading to better adaptation to complex dynamics. This effect is particularly evident in with multi-phase control tasks (e.g. in the walker domain), where different behavior modes enable a more efficient division of the action space. Also, the multimodal policy exhibits greater stability, as reflected in its reduced variance across different seeds. This suggests that the structured nature of discrete action modes provides a more consistent and reliable learning process, reducing the risk of poor policy initialization and early convergence to a suboptimal policy.

\subsection{Sampling Strategy}
Figure \ref{fig:exp_0} additionally compares the performance of our method using two different discrete sampling strategies: Gumbel-Softmax reparameterization and STE. Our results consistently show that the STE approach exhibits better stability and performance in all six tasks.

The advantage of STE is attributed to its ability to retain the discrete nature of the sampled variables during the forward pass, while still allowing gradient-based optimization via a simple and computationally efficient backward pass. We hypothesize that the additional sampling step involved with Gumbel-Softmax reparameterization introduces higher gradient variance that leads to instability during optimization.

\begin{figure}[h]
    \centering
    \includegraphics[width=0.99\linewidth]{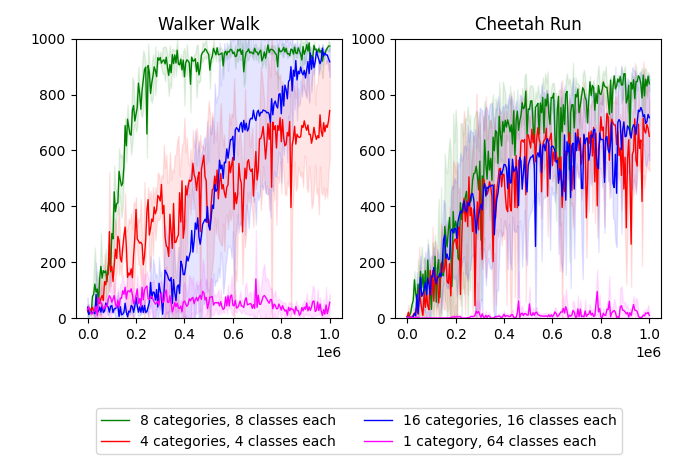}
    % \captionsetup{justification=centerlast, singlelinecheck=false}
    \caption{Comparison of design choice for behavior modes and justification of multiple categorical variables usage. Plot shows test performance over environment steps. The line shows the average and the shaded area represents 1 standard deviation over 5 seeds.}
    \label{fig:exp_1}
\end{figure}

\subsection{Design Choice for Behavior Modes}
Finally, Figure \ref{fig:exp_1} demonstrates the importance of selecting an appropriate number of discrete behavior modes and justifies our choice of using multiple categorical variables. We experiment with four different settings, such as $(N, M = 4)$, $(N, M = 8)$, $(N, M = 16)$ and $(N =1,  M= 64)$, to encode the intermediate behavior mode with categorical variables. The results show that having too few behavior modes limits the expressiveness of the policy, reducing its ability to effectively explore and adapt to complex control tasks. Conversely, an excessively large number of modes can lead to slower learning, as the policy struggles to allocate meaningful gradients across a high-dimensional discrete space. 

Note that, for the single categorical variable case $(N =1,  M= 64)$, even with the same behavior mode dimension ($8 \times 8$ vs. $1 \times 64$) the agent does not learn any meaningful behavior in both tasks. This provides evidence to the significance of multiple categorical variables over just one. By incorporating multiple categorical variables, our approach strikes a balance between flexibility and efficiency. Instead of relying on a single categorical variable with an impractically large number of classes, using multiple categorical variables with fewer classes each introduces a structured way to represent diverse behaviors without overwhelming the optimization process. This structured representation allows the policy to capture hierarchical patterns in action selection, leading to more stable and effective learning.

\section{CONCLUSIONS}
\label{concl}
We present Categorical Policy, a multimodal approach for continuous control that introduces discrete behavior modes to enhance exploration and adaptability. Unlike traditional unimodal policies that rely solely on a single action distribution, our approach first samples a discrete action mode before generating fine-grained continuous actions. We overcome the challenges of using a single categorical variable, which would require an impractically large number of modes to achieve fine control, by employing multiple categorical variables with fewer classes each. Empirical results show that this structured representation leads to a more expressive policy capable of structured exploration and faster convergence, thus consistently matching or outperforming traditional unimodal policies across a variety of continuous control tasks. We further show that choosing the appropriate sampling scheme plays a crucial role in training stability and performance. Specifically, we compare Gumbel-Softmax reparameterization and Straight-Through Estimation (STE), and find that STE yields slightly more stable training due to its simpler, noise-free gradient path.

 % TODOS
 % 1. Add figures
 % 2. Add figure captions


%%%%%%%%%%%%%%%%%%%%%%%%%%%%%%%%%%%%%%%%%%%%%%%%%%%%%%%%%%%%%%%%%%%%%%%%%%%%%%%%

%%%%%%%%%%%%%%%%%%%%%%%%%%%%%%%%%%%%%%%%%%%%%%%%%%%%%%%%%%%%%%%%%%%%%%%%%%%%%%%%

%%%%%%%%%%%%%%%%%%%%%%%%%%%%%%%%%%%%%%%%%%%%%%%%%%%%%%%%%%%%%%%%%%%%%%%%%%%%%%%%

%%%%%%%%%%%%%%%%%%%%%%%%%%%%%%%%%%%%%%%%%%%%%%%%%%%%%%%%%%%%%%%%%%%%%%%%%%%%%%%%

\bibliographystyle{IEEEtran}
\bibliography{ref}

\end{document}